\documentclass[runningheads,a4paper]{llncs}
\usepackage{amssymb}
\usepackage{graphicx}
\usepackage{url}
\urldef{\mailsa}\path|art@tx.technion.ac.il|
\urldef{\mailsb}\path|bron@eng.tau.ac.il|
\urldef{\mailsc}\path|michael.bronstein@usi.ch|
\urldef{\mailsd}\path|ron@cs.technion.ac.il|

\usepackage{float}
\usepackage{subfigure}
\usepackage{graphicx}
\usepackage{listings}
\DeclareGraphicsExtensions{.jpg,.eps}
\usepackage{amsmath}
\usepackage{amsfonts}
\usepackage{amssymb}
\usepackage{enumerate}
\usepackage{textcomp}
\usepackage{color}
\usepackage{algorithmic}
\usepackage{algorithm}

\newcommand{\RR}{{\mathbb{R}}}

 {\everymath{\displaystyle\everymath{}}\array}%
 {\endarray}

\begin{document}

\mainmatter

\title{Diffusion framework for geometric and photometric data fusion in non-rigid shape analysis}

\titlerunning{Framework for geometric and photometric data fusion}
\authorrunning{A. Kovnatsky {\em et al.}}

\author{
Artiom Kovnatsky$^1$ \and Michael M. Bronstein$^3$ \and Alexander M. Bronstein$^4$ \and \\ Ron Kimmel$^2$
}

\institute{
$^1$Department of Mathematics, \\
\mailsa\\
$^2$Department of Computer Science,\\
\mailsd\\ 
Technion, Israel Institute of Technology, Haifa, Israel\vspace{1mm}\\
$^2$Inst. of Computational Science, Faculty of Informatics,\\ Universit{\`a} della Svizzera Italiana, Lugano, Switzerland\\
\mailsc\vspace{1mm}\\
$^3$Dept. of Electrical Engineering, Tel Aviv University, Israel\\
\mailsb
}
\maketitle

\begin{abstract}
In this paper, we explore the use of the diffusion geometry framework for the fusion of geometric and photometric information in local and global shape descriptors.
Our construction is based on the definition of a diffusion process on the shape manifold embedded into a high-dimensional space where the embedding coordinates represent the photometric information.
Experimental results show that such data fusion is useful in coping with different challenges of shape analysis where pure geometric and pure photometric methods fail.\end{abstract}

\section{Introduction}
\label{sec:intro}

\begin{figure}[t]
    \begin{center}
    \includegraphics*[width=1\linewidth,bb=1 1 986 619,clip]{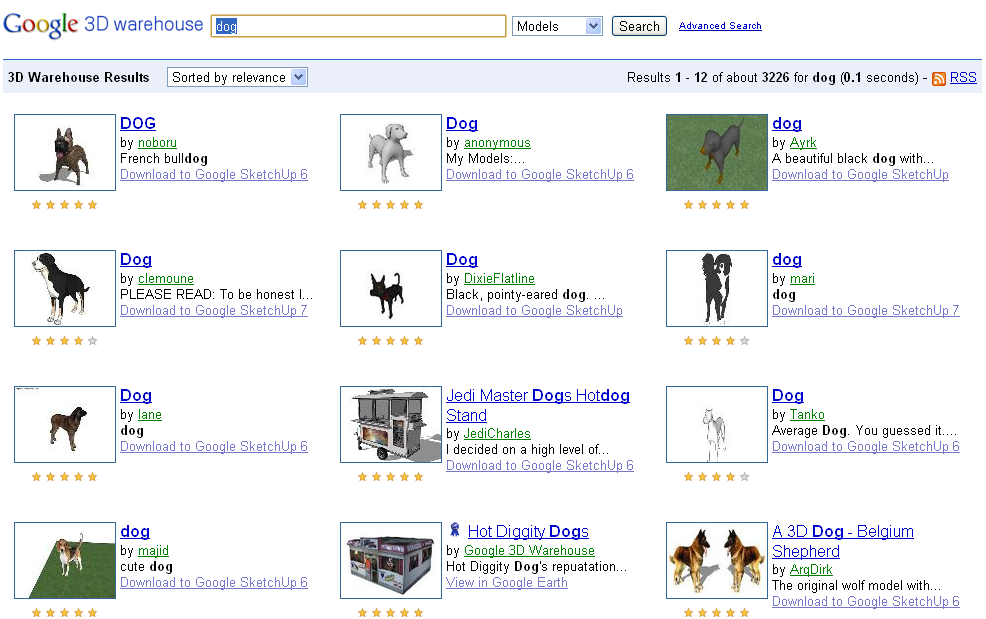}
   \caption{ \small  The need for content-based shape retrieval: text-based search engine such as \textit{Google 3D Warehouse} returns shapes of dogs as well as hot-dog cabins in response to the query ``dog''. The later is obviously irrelevant. \label{fig:googles}}
    \end{center}
\end{figure}

In last decade, the amount of geometric data available in the public domain, such as Google 3D Warehouse, has grown dramatically and created the demand for shape search and retrieval algorithms capable of finding similar shapes in the same way a search engine responds to text queries.
However, while text search methods are sufficiently developed to be ubiquitously used, the search and retrieval of 3D shapes remains a challenging problem. Shape retrieval based on text metadata, like annotations and tags added by the users, is often incapable of providing relevance level required for a reasonable user experience (see Figure \ref{fig:googles}).

\textit{Content-based shape retrieval} using the shape itself as a query and based on the comparison of 
geometric and topological properties of shapes is complicated by the fact that many 3D objects manifest rich variability, and shape retrieval must often be \textit{invariant} under different classes of transformations. A particularly challenging setting is the case of non-rigid shapes, including a wide range of transformations such as bending and articulated motion, rotation and translation, scaling, non-rigid deformation, and topological changes.
The main challenge in shape retrieval algorithms is computing a \textit{shape descriptor}, that would be unique for each shape, simple to compute and store, and invariant under different type of transformations. Shape similarity is determined by comparing the shape descriptors.

{\bf Prior works. }
Broadly, shape descriptors can be divided into {\em global} and {\em local}. The former consider global geometric or topological shape characteristics such as distance distributions \cite{osada2002sd,rustamov2007lbe,MS_GMOD}, geometric moments \cite{moments:kazhdan2003ris,moments:vranic2001tor}, or spectra \cite{reuter}, whereas the latter describe the local behavior of the shape in a small patch.
Popular examples of local descriptors include spin images \cite{assfalg:bert:bimbo:pala:spinimage}, 
shape contexts \cite{amores2007context}, integral volume descriptors \cite{gelfand2005rgr} and radius-normal histograms \cite{pan:zhang:zhang:ye:3Dslices}.
Using the bag of features paradigm common in image analysis \cite{siv:zis:CVPR:03,chum:ICCV:07}, 
a global shape descriptor counting the occurrence of local descriptors in some vocabulary can be computed \cite{BroBroOvsGui09}.

Recently, there has been an increased interest in the use of {\em diffusion geometry} \cite{diff,levy2006lbe} for constructing invariant shape descriptors.
Diffusion geometry is closely related to heat propagation properties of shapes and allows obtaining global descriptors, such as 
%
distance distributions \cite{rustamov2007lbe,MS_GMOD,BroBB1} and Laplace-Beltrami spectral signatures \cite{reuter}, as well local descriptors such as 
heat kernel signatures \cite{sunHKS,BroSI:HKS}.
%
In particular, heat kernel signatures \cite{sunHKS} showed very promising results in large-scale shape retrieval applications \cite{BroBroOvsGui09}.

One limitation of these methods is that, so far, only {\em geometric} information has been considered. However, the abundance of textured models in computer graphics and modeling applications, as well as the advance in 3D shape acquisition 
\cite{YPS10,ZBH07} 
allowing to obtain textured 3D shapes of even moving objects, bring forth the need for descriptors also taking into consideration \textit{photometric} information. 
Photometric information plays an important role in a variety of shape analysis applications, such as shape matching and correspondence \cite{thorstensen,wyngaerd}.
Considering 2D views of the 3D shape \cite{wu:VIP,ohb:osad:fur:ban:salfeat}, standard feature detectors and descriptors used in image analysis such as SIFT \cite{low:IJCV:04} can be employed.
More recently, Zaharescu \textit{et al.} \cite{zaharescu-surface} proposed a geometric SIFT-like descriptor for textured shapes, defined directly on the surface.

{\bf Main contribution. }
In this paper, we extend the diffusion geometry framework to include photometric information in addition to its geometric counterpart.
This way, we incorporate important photometric properties on one hand, while exploiting a principled and theoretically established approach on the other.
The main idea is to define a diffusion process that takes into consideration not only the geometry but also the texture of the shape. This is achieved by considering the shape as a manifold  in a higher dimensional combined geometric-photometric embedding space, similarly to methods in image processing applications \cite{kimmel2000images,Ling05deformationinvariant}. As a result, we are able to construct local descriptors (heat kernel signatures) and global descriptors (diffusion distance distributions).
The proposed data fusion can be useful in coping with different challenges of shape analysis where pure geometric and pure photometric methods fail. 



\section{Background}
\label{sec:backgr}

Throughout the paper, we assume the shape to be modeled as a two-dimensional compact Riemannian manifold $X$ (possibly with a boundary)
equipped with a metric tensor $g$.
%
%
Fixing a system of local coordinates on $X$, the latter can be expressed as a $2\times 2$ matrix $g_{\mu\nu}$, also known as the first fundamental form.
The metric tensor allows to express the length of a vector $v$ in the tangent space $T_x X$ at a point $x$ as $g_{\mu\nu} v^\mu v^\nu$, where repeated indices $\mu,\nu = 1,2$ are summed over following Einstein's convention.

Given a smooth scalar field $f : X \rightarrow \mathbb{R}$ on the manifold, its {\em gradient} is defined as the vector field $\nabla f$ satisfying $f(x+dx) = f(x) + g_x( \nabla f(x), dx )$ for every point $x$
and every infinitesimal tangent vector  $dx \in T_x X$.
The metric tensor $g$ defines the {\em Laplace-Beltrami operator} $\Delta_g$ that satisfies
\begin{eqnarray}
\int f \Delta_g h \, da &=& - \int g_x( \nabla f, \nabla h ) da
\label{eq:lbo}
\end{eqnarray}
for any pair of smooth scalar fields $f,h:X\rightarrow \mathbb{R}$; here $da$ denotes integration with respect to the standard area measure on $X$. Such an integral definition is usually known as the Stokes identity.
The Laplace-Beltrami operator is positive semi-definite and self-adjoint. Furthermore, it is
an \textit{intrinsic} property of $X$, i.e., it is expressible solely in terms of $g$.
In the case when the metric $g$ is Euclidean, $\Delta_g$ becomes the standard Laplacian.

The Laplace-Beltrami operator gives rise to the \textit{heat equation},
\begin{equation}
\label{eq:heat}
\left (\Delta_g + \frac{\partial }{\partial t}\right) u = 0,
\end{equation}
which describes diffusion processes and heat propagation on the manifold. Here, $u(x,t)$ denotes the distribution of heat at time $t$ at point $x$.
The initial condition to the equation is some heat distribution $u(x,0)$, and if the manifold has a boundary, appropriate boundary conditions
(e.g. Neumann or Dirichlet) must be specified.
The solution of~(\ref{eq:heat}) with a point initial heat distribution ${u_0}\left( x \right) = \delta \left( {x - x'} \right)$ is called the \textit{heat kernel} and denoted here by $h_t(x,x')$.
Using a signal processing analogy, $h_t$ can be thought of as the ``impulse response'' of the heat equation.

By the spectral decomposition theorem, the heat kernel can be represented as \cite{jones2008mpe}
\begin{equation}
\label{eq:heatkernel1}
h_t(x,x') = \sum_{i \ge 0} e^{-\lambda_i t}\phi_i(x) \phi_i(x),
\end{equation}
where $0=\lambda_0 \leq \lambda_1 \leq \hdots $ are the eigenvalues and $\phi_0,\phi_1,\hdots$ the corresponding eigenfunctions of the
Laplace-Beltrami operator (i.e., solutions to $\Delta_g \phi_i = \lambda_i\phi_i$).
The value of the heat kernel $h_t(x,x')$ can be interpreted as the transition probability
density of a random walk of length $t$ from the point $x$ to the point $x'$.
This allows to construct a family of intrinsic metrics known as \emph{diffusion metrics},\footnote{Note that here the term \emph{metric} is understood in the sense
of metric geometry rather than the Riemannian inner product. To avoid confusion, we refer to the latter as to \emph{metric tensor} throughout the paper.}
\begin{eqnarray}
d^2_t(x,x') &=& \int \left( h_t(x,\cdot) - h_t(x',\cdot) \right)^2 da \nonumber \\
&=& \sum_{i> 0} e^{-\lambda_i t} (\phi_i(x) - \phi_i(x'))^2.
\label{eq:diffdist}
\end{eqnarray}
These metrics have an inherent multi-scale structure and measure the ``connectivity rate'' of the two points by paths of length $t$.
We will collectively refer to quantities expressed in terms of the heat kernel or diffusion metrics as to \emph{diffusion geometry}.
Since the Laplace-Beltrami operator is intrinsic, the diffusion geometry it induces is invariant under isometric deformations of $X$ (incongruent embeddings of $g$ into $\RR^3$).

\section{Fusion of geometric and photometric data}
\label{sec:chks}

Let us further assume that the Riemannian manifold $X$ is a submanifold of some manifold $\mathcal{E}$ ($\mathrm{dim}(\mathcal{E})=m>2$) with 
the Riemannian metric tensor $h$, embedded by means of a diffeomorphism $\xi: X \rightarrow \mathcal{E}$.
A Riemannian metric tensor on $X$ induced by the embedding is the {\em pullback metric} $(\xi^*h)(r,s) = h(d\xi(r),d\xi(s))$ for $r,s\in T_x X$, where $d\xi:T_x X \rightarrow T_{\xi(x)}\mathcal{E}$ is the differential of $\xi$.
In coordinate notation, the pullback metric is expressed as $(\xi^*h)_{\mu\nu} = h_{ij}\partial_\mu \xi^i \partial_\nu \xi^j$, where the indices $i,j=1,\hdots,m$ denote the embedding coordinates.

Here, we use the structure of $\mathcal{E}$ to model joint geometric and photometric information. Such an approach has been successfully used in image processing \cite{kimmel2000images}.
When considering shapes as geometric object only, we define $\mathcal{E} = \RR^3$ and $h$ to be the Euclidean metric. In this case, $\xi$ acts as a {\em parametrization} of $X$ and the pullback metric becomes simply $(\xi^*h)_{\mu\nu}=\partial_\mu \xi^1 \partial_\nu \xi^1 + \hdots + \partial_\mu \xi^3 \partial_\nu \xi^3 = \langle \partial_\mu \xi, \partial_\nu \xi \rangle_{\RR^3}$.
In the case considered in this paper, the shape is endowed with photometric information given in the form of a field $\alpha:X\rightarrow \mathcal{C}$, where $\mathcal{C}$ denotes some colorspace (e.g., RGB or Lab). This photometric information can be modeled by defining $\mathcal{E} = \RR^3\times \mathcal{C}$ and an embedding $\xi = (\xi_g, \xi_p)$. The embedding coordinates corresponding to geometric information $\xi_g = (\xi^1,\hdots,\xi^3)$ are as previously and the embedding coordinate corresponding to photometric information are given by
$\xi_p = (\xi^{4},\hdots,\xi^6)=\eta(\alpha^1,\hdots,\alpha^3)$, where $\eta \geq 0$ is a scaling constant.
Simplifying further, we assume $\mathcal{C}$ to have a Euclidean structure (for example, the Lab colorspace has a natural Euclidean metric). The metric in this case boils down to $(\xi^*h)_{\mu\nu}=\langle \partial_\mu \xi_g, \partial_\nu \xi_g \rangle_{\RR^3} + \eta^2 \langle \partial_\mu \xi_p, \partial_\nu \xi_p \rangle_{\RR^3}$, which hereinafter we shall denote by $\hat{g}_{\mu\nu}$.

\begin{figure}[t]
   \begin{center}
\begin{minipage}[b]{0.32\linewidth}
  \centering   \includegraphics*[width=0.65\linewidth,bb=1 1 240 547]{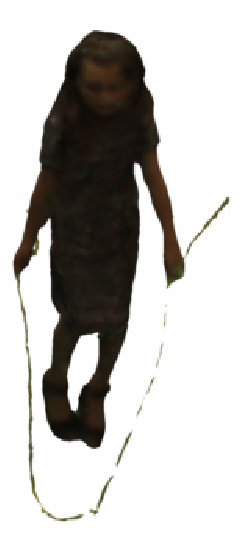}
\end{minipage}
\begin{minipage}[b]{0.32\linewidth}
  \centering     \includegraphics*[width=0.65\linewidth,bb=1 1 228 535]{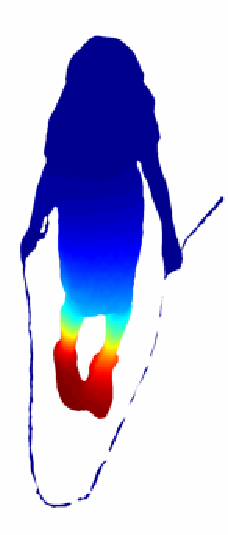}
\end{minipage}
\begin{minipage}[b]{0.32\linewidth}
  \centering     \includegraphics*[width=0.65\linewidth,bb=1 1 228 535]{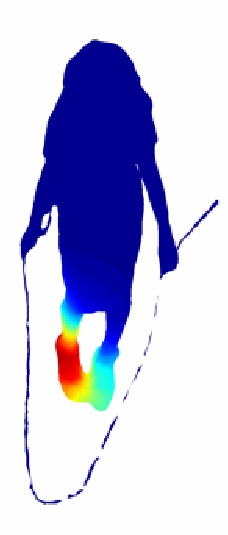}
\end{minipage}
   \end{center}
  \label{fig:diff}
  \caption{ \label{fig:fus1} \small Textured shape (left); values of the heat kernel ($x$ placed on the foot, $t=1024$) arising from regular purely geometric (middle) and mixed photometric-geometric (right) diffusion process.}
\end{figure}

The Laplace-Beltrami operator $\Delta_{\hat{g}}$ associated with such a metric gives rise to diffusion geometry that combines photometric and geometric information (Figure~\ref{fig:fus1}).

{\bf Invariance. }
It is important to mention that the joint metric tensor $\hat{g}$ and the diffusion geometry it induces have inherent ambiguities.
Let us denote by $\mathrm{Iso}_g = \mathrm{Iso}((\xi_g^*h)_{\mu\nu})$ and
$\mathrm{Iso}_p = \mathrm{Iso}((\xi_p^*h)_{\mu\nu})$
the respective groups of transformation that leave the geometric and the photometric components of the shape unchanged.
We will refer to such transformations as geometric and photometric isometries.
The diffusion metric induced by $\hat{g}$ is invariant the joint isometry group $\mathrm{Iso}_{\hat{g}} = \mathrm{Iso}((\xi_g^*h)_{\mu\nu} + \eta^2(\xi_p^*h)_{\mu\nu})$.
Ideally, we would like $\mathrm{Iso}_{\hat{g}} = \mathrm{Iso}_g \times \mathrm{Iso}_p$ to hold. In practice, $\mathrm{Iso}_{\hat{g}}$ is bigger: while every composition
of a geometric isometry with a photometric isometry is a joint isometry,
there exist some joint isometries which cannot be obtained as a composition of geometric and photometric isometries. An example of such transformations is uniform scaling of $(\xi_g^*h)_{\mu\nu}$ combined with compensating scaling of $(\xi_p^*h)_{\mu\nu}$.
The ambiguity stems from the fact that $\mathrm{Iso}_{\hat{g}}$ is bigger compared to $\mathrm{Iso}_g \times \mathrm{Iso}_p$.
%
Experimental results show that no realistic geometric and photometric transformations lie in $\mathrm{Iso}_{\hat{g}} \setminus (\mathrm{Iso}_g \times \mathrm{Iso}_p)$, however, a formal  characterization of the isometry group is an important theoretical question for future research.

\section{Numerical implementation}
\label{sec:num}


Let $\{x_1,\hdots,x_N\} \subseteq X$ denote the discrete samples of the shape, and $\xi(x_1),\hdots,\xi(x_N)$ be the corresponding embedding coordinates (three-dimensional in the case we consider only geometry, or six-dimensional in the case of geometry-photometry fusion).
We further assume to be given a {\em triangulation} (simplicial complex), consisting of {\em edges} $(i,j)$ and {\em faces} $(i,j,k)$ where each $(i,j), (j,k)$, and $(i,k)$ is an edge (here $i,j,k=1,\hdots,N$).

{\bf Discrete Laplacian. }
A function $f$ on the discretized manifold is represented as an $N$-dimensional vector $(f(x_1),\hdots,f(x_N))$. The discrete Laplace-Beltrami operator can be written in the generic form
\begin{equation}
\label{eq:lplbeltrdiscrgen}
(\hat{\Delta} f)(x_i) = \frac{1}{a_i} \sum_{j \in \mathcal{N}_i} w_{ij} (f(x_i) - f(x_j)),
\end{equation}
where $w_{ij}$ are weights, $a_i$ are normalization coefficients, and $\mathcal{N}_i$ denotes a local neighborhood of point $i$.
Different discretizations of the Laplace-Beltrami operator can be cast into this form by appropriate definition of the above constants.
%
%
For shapes represented as triangular meshes, a widely-used method is the {\em cotangent scheme},
%
which preserves many important properties of the continuous Laplace-Beltrami operator, such as positive semi-definiteness, symmetry, and locality \cite{wardetzky2008discrete}.
Yet, in general, the cotangent scheme does not converge to the continuous Laplace-Beltrami operator, in the sense that the solution of the discrete eigenproblem does not converge to the continuous one (pointwise convergence exists if the triangulation and sampling satisfy certain conditions \cite{xu:04:LB}).

Belkin \textit{et al.} \cite{BelSunWanSODA09} proposed a discretization which is convergent without the restrictions on ``good'' triangulation required by the cotangent scheme.
In this scheme, $\mathcal{N}_i$ is chosen to be the entire sampling $\{x_1,\hdots,x_N\}$, ${a_i} = \frac{1}{{4\pi \rho^2} }$, and $w_{ij} = S_j e^{- \| \xi(x_i) -\xi(x_j) \|^2 / 4\rho}$, where $\rho$ is a parameter.
In the case of a Euclidean colorspace, $w_{ij}$ can be written explicitly as
\begin{equation}
w_{ij} = S_j \exp \left\{- \frac{\|\xi_g(x_i) - \xi_g(x_j)\|^2}{4\rho} - \frac{\|\xi_p(x_i) - \xi_p(x_j)\|^2}{4\sigma}\right\}
\end{equation}
where $\sigma = \rho/\eta^2$, which resembles the weights used in the {\em bilateral filter} 
\cite{bilatfilt}.
%
%
Experimental results also show that this operator produces accurate approximation of the Laplace-Beltrami operator under various conditions, such as noisy data input and different sampling \cite{than:ms,BelSunWanSODA09}.



{\bf Heat kernel computation. }
In matrix notation, equation~(\ref{eq:lplbeltrdiscrgen}) can be written as
%
$\hat{\Delta} f = A^{-1} W f$,
%
where $A = \mathrm{diag}(a_i)$ and $W = \mathrm{diag}\left(\sum_{l\neq i} w_{il} \right) - (w_{ij})$. 
The eigenvalue problem $\hat{\Delta} \Phi = \Lambda \Phi$ is equivalent to the generalized symmetric eigenvalue problem
$W \Phi = \Lambda A \Phi$,
%
where $\Lambda = \mathrm{diag}(\lambda_0,\hdots,\lambda_K)$ is the diagonal matrix of the first $K$ eigenvalues, and $\Phi = (\phi_0, \hdots, \phi_K)$ is the matrix of the eigenvectors stacked as columns.
Since typically $W$ is sparse, this problem can be efficiently solved numerically.

Heat kernels can be approximated by taking the first largest eigenvalues and the corresponding eigenfunctions in~(\ref{eq:heatkernel1}). Since the coefficients in the expansion of $h_t$ decay as $\mathcal{O}(e^{-t})$, typically a few eigenvalues ($K$ in the range of $10$ to $100$) are required.


%

\section{Results and applications}
\label{sec:res}

In this section, we show the application of the proposed framework to retrieval of textured shapes. We compare two approaches: bags of local features and distributions of diffusion distances.

\subsection{Bags of local features}


{\bf ShapeGoogle framework. }
Sun \emph{et al.} \cite{sunHKS} proposed using the heat propagation properties as a local descriptor of the manifold.
The diagonal of the heat kernel,
\begin{eqnarray}
h_t(x,x) &=& \sum_{i \ge 0} e^{-\lambda_i t}\phi_i^2(x),
\end{eqnarray}
referred to as the {\em heat kernel signature} (HKS), captures the local properties of $X$ at point $x$ and scale $t$.
The descriptor is computed at each point as a vector of the values $p(x) = (K_{t_1}(x,x),\hdots,K_{t_n}(x,x))$, where $t_1,\hdots,t_n$ are some time values.
Such a descriptor is deformation-invariant, easy to compute, and provably informative \cite{sunHKS}.
%


Ovsjanikov \emph{et al.} \cite{BroBroOvsGui09} employed the HKS local descriptor for large-scale shape retrieval using the {\em bags of features} paradigm \cite{siv:zis:CVPR:03}.
In this approach, the shape is considered as a collection of ``geometric words'' from a fixed ``vocabulary'' and is described by the distribution of such words, also referred to as a \textit{bag of features} or BoF.
The vocabulary is constructed offline by clustering the HKS descriptor space. Then, for each point on the shape, the HKS is replaced by the nearest vocabulary word by means of vector quantization. Counting the frequency of each word, a BoF is constructed.
The similarity of two shapes $X$ and $Y$ is then computed as the distance between the corresponding BoFs, $d(X,Y) = \| \mathrm{BoF}_X - \mathrm{BoF}_Y \|$.

Using the proposed approach, we define the {\em color heat kernel signature} (cHKS), defined in the same way as HKS with the standard Laplace-Belrami operator replaced by the one resulting from the geometric-photometric embedding. In the following, we show that such descriptors allow achieving superior retrieval performance.

%

\begin{figure}[ht]
\begin{center}
  \includegraphics*[width=1\linewidth]{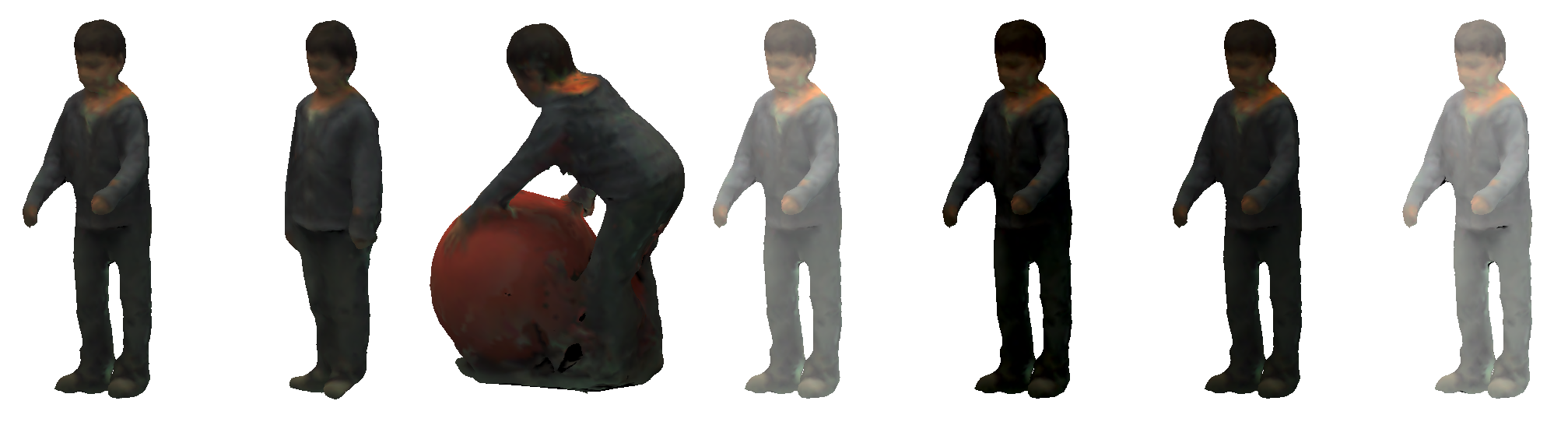} \\
  \includegraphics*[width=0.75\linewidth]{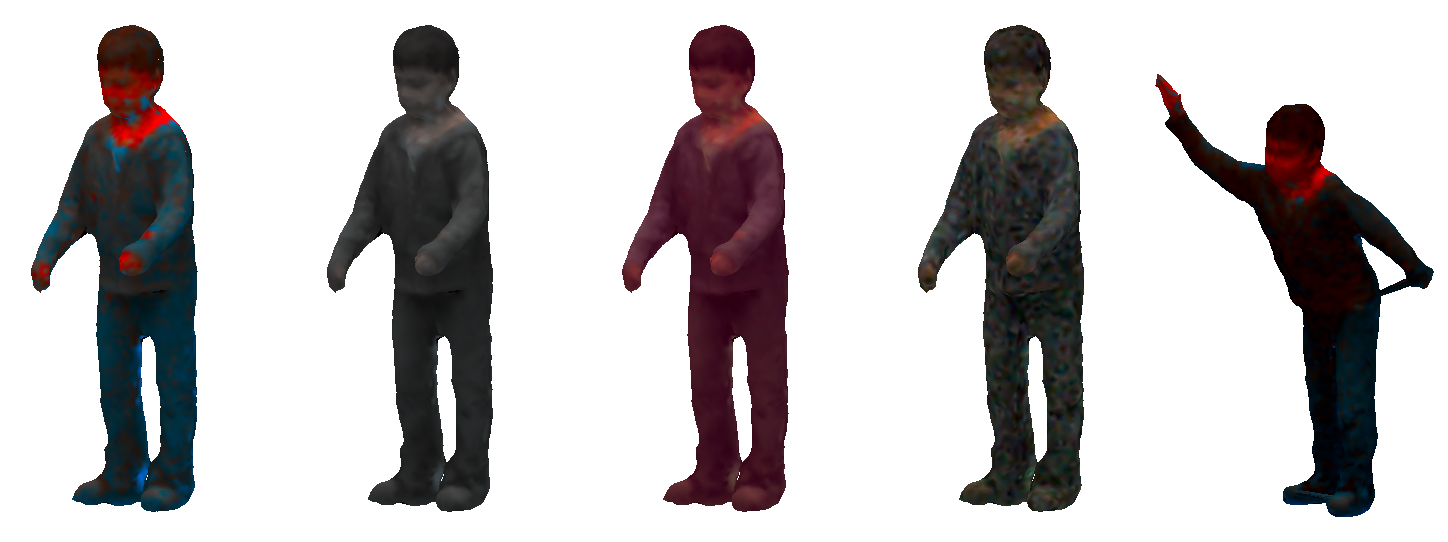} 
\end{center}
  \caption{ \label{fig:trnsf:shapes} \small Examples of geometric and photometric shape transformations used as queries (shown at strength 5).
  First row, left to right: null, isometry+topology, partiality, two brightness transformations (brighten and darken), two contrast transformations (increase and decrease contrast). Second row, left to right: two saturation transformations (saturate and desaturate), hue, color noise, mixed.}
\end{figure}

\begin{figure}[!ht]
 \centering
 \includegraphics*[width=1\linewidth]{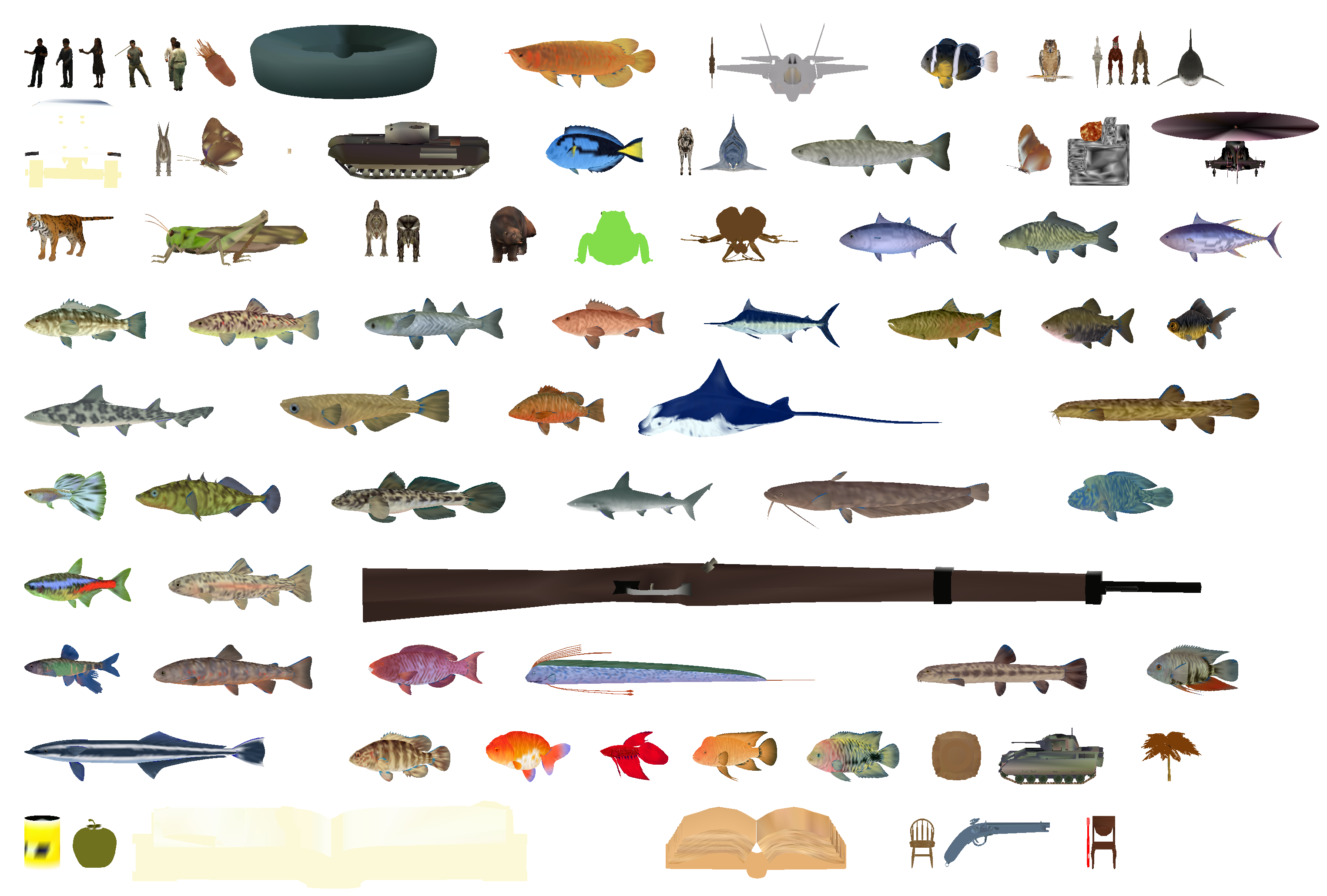}
  \caption{ \label{fig:data:shapes} \small Null shapes in the dataset (shown at arbitrary scale for visualization purposes).}
\end{figure}

{\bf Evaluation methodology. }
In order to evaluate the proposed method, we used the SHREC 2010 robust large-scale shape retrieval benchmark methodology \cite{SHRECr}.
The query set consisted of 270 real-world human shapes from 5 classes acquired by a 3D scanner with real geometric
transformations and simulated photometric transformations of different types
and strengths, totalling in 54 instances per shape (Figure~\ref{fig:trnsf:shapes}).
Geometric transformations were divided into {\em isometry+topology} (real articulations and topological changes due to acquisition imperfections), and
{\em partiality} (occlusions and addition of clutter such as the red ball in Figure~\ref{fig:trnsf:shapes}).
Photometric transformations included {\em contrast} (increase and decrease by scaling of the $L$ channel), {\em brightness} (brighten and darken by shift of the $L$ channel), {\em hue} (shift in the $a$ channel), {\em saturation} (saturation and desaturation by scaling of the $a,b$ channels), and {\em color noise} (additive Gaussian noise in all channels).
{\em Mixed} transformations included isometry+topology transformations in combination with two randomly selected photometric transformations.
In each class, the transformation appeared in five different versions numbered 1--5 corresponding to the transformation strength levels.
One shape of each of the five classes was added to the queried corpus
in addition to other 75 shapes used as clutter (Figure~\ref{fig:data:shapes}).
%

Retrieval was performed by matching 270 transformed queries to the 75 null shapes. Each query had exactly one correct corresponding null shape in the dataset.
Performance was evaluated using the precision-recall characteristic.
{\em Precision} $P(r)$ is defined as the percentage of relevant shapes in the first $r$ top-ranked retrieved shapes.
{\em Mean average precision} (mAP), defined as $mAP = \sum_r P(r) \cdot rel(r)$,
where $rel(r)$ is the relevance of a given rank, was used as a single measure of performance. Intuitively, mAP is interpreted as the area below the precision-recall curve.
Ideal retrieval performance results in first relevant match with mAP=100\%.
Performance results were broken down according to transformation class and strength.

%

{\bf Methods. }
In additional to the proposed approach, we compared purely geometric,
purely photometric, and joint photometric-geometric descriptors.
As a purely geometric descriptor, we used bags of features based on HKS according to \cite{BroBroOvsGui09};
purely photometric shape descriptor was a color histogram. As joint
photometric-geometric descriptors, we used bags of features computed
with the MeshHOG \cite{zaharescu-surface} and the proposed color HKS
(cHKS).

For the computation of the bag of features descriptors, we used the
Shape Google framework with most of the settings as proposed in
\cite{BroBroOvsGui09}.
More specifically, HKS were computed at six scales ($t = 1024, 1351.2,
1782.9, 2352.5$, and $4096$). Soft vector quantization was applied
with variance taken as twice the median of all distances between
cluster centers. Approximate nearest neighbor method
\cite{AryaMount:98:ICP} was used for vector quantization.
The Laplace-Beltrami operator discretization was computed using the
Mesh-Laplace scheme \cite{belkin2009constructing} with scale parameter
$\rho=2$.
Heat kernels were approximated using the first $200$ eigenpairs of the
discrete Laplacian.
The MeshHOG descriptor was computed at prominent feature points
(typically 100-2000 per shape), detected using the MeshDOG detector
\cite{zaharescu-surface}.
The vocabulary size in all the cases was set to $48$.

In cHKS, in order to avoid the choice of an arbitrary value $\eta$, we
used a set of three different weights ($\eta=0,0.05,0.1$) to compute
the cHKS and the corresponding BoFs.
The distance between two shapes was computed as the sum of the
distances between the corresponding BoFs for each $\eta$, weighted by
$\eta$, and 1 in case of $\eta=0$,
$d(X, Y)= \| \mathrm{BoF}_{X}^{0}-\mathrm{BoF}_{Y}^{0}  \|^2_{1} +
\sum_{\eta} \eta \| \mathrm{BoF}_{X}^{\eta}-\mathrm{BoF}_{Y}^{\eta}
\|^2_{1}$.

\begin{figure*}[tpb]
\begin{center}
\begin{minipage}[b]{0.3768\linewidth}
  \centering \includegraphics*[width=1\columnwidth]{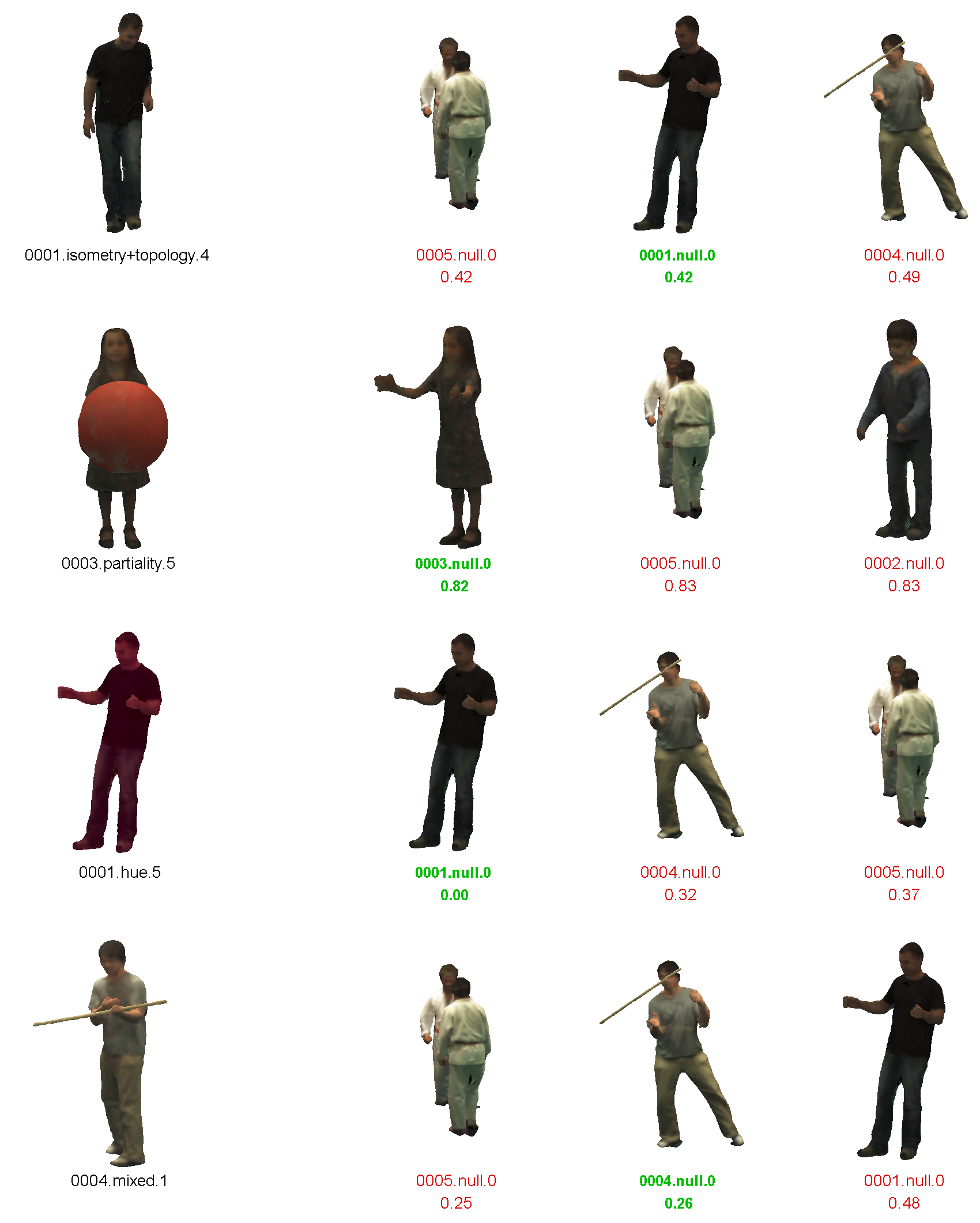}
    \centerline{\hspace{12mm}\footnotesize{HKS BoF \cite{BroBroOvsGui09}}}
\end{minipage}\hspace{3mm}
\begin{minipage}[b]{0.2837\linewidth}
 \centering \includegraphics*[width=1\columnwidth]{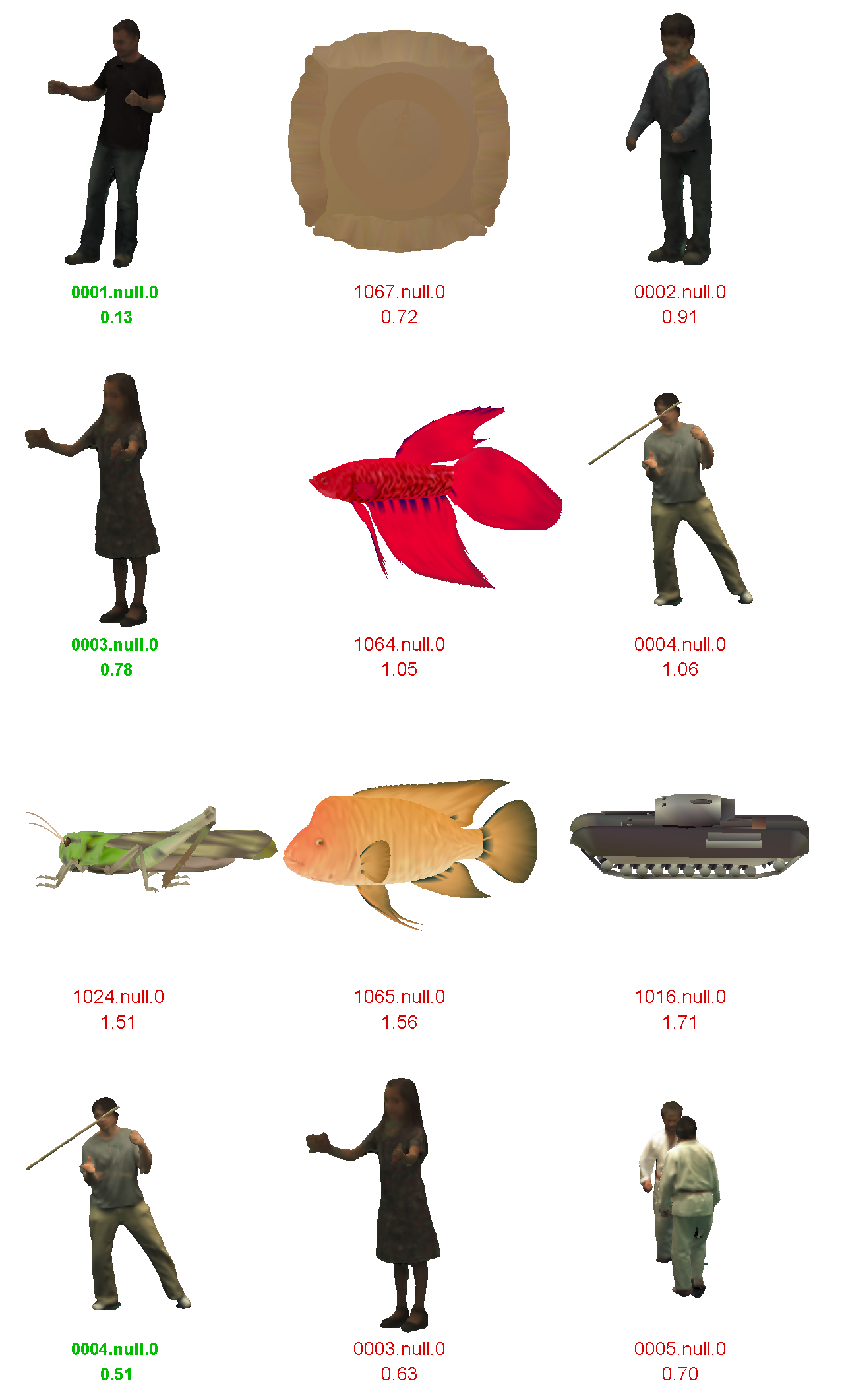}
    \centerline{\hspace{0mm}\footnotesize{Color histogram}}
\end{minipage}\hspace{3mm}
\begin{minipage}[b]{0.2394\linewidth}
 \centering \includegraphics*[width=1\columnwidth]{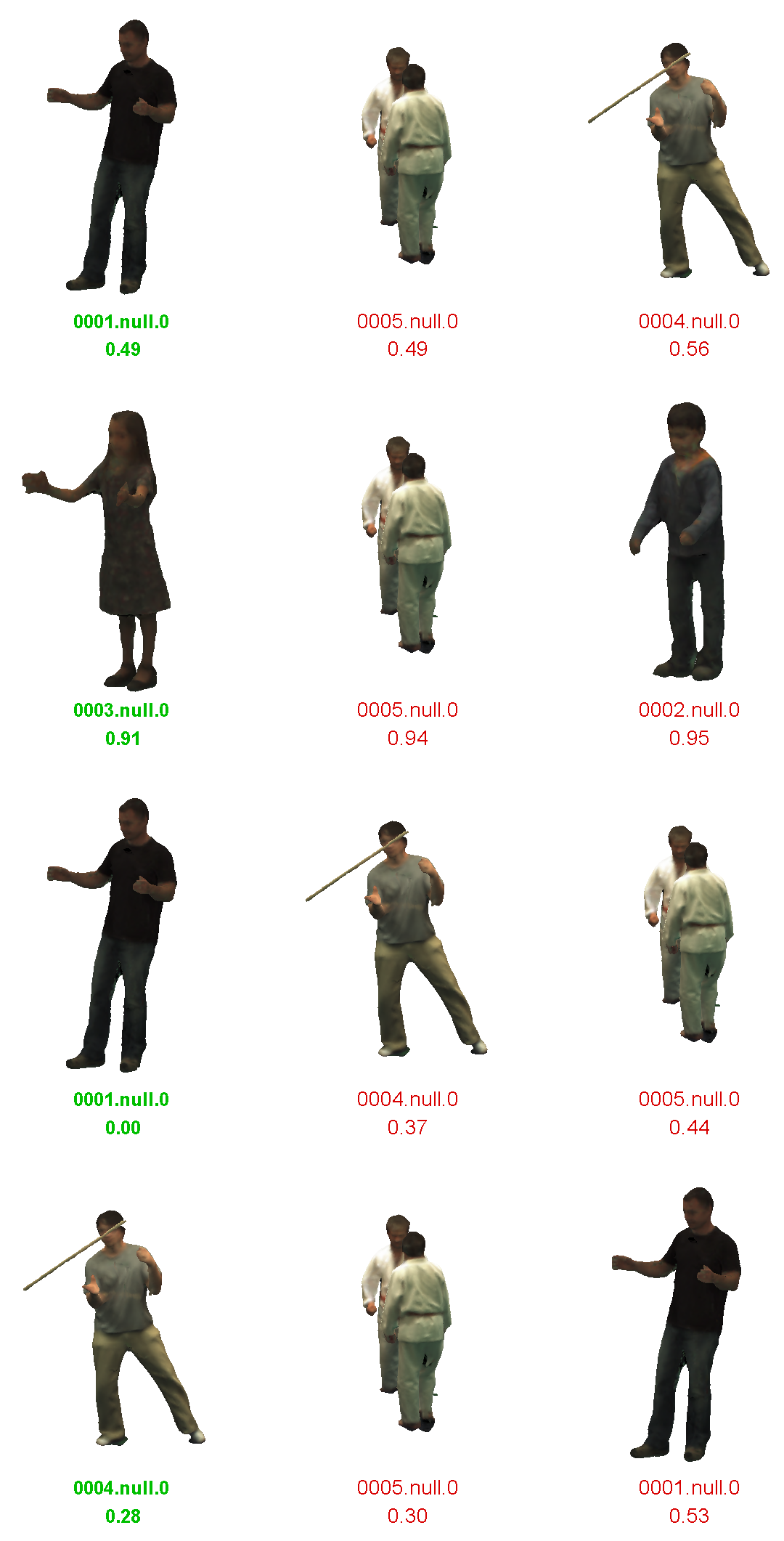}
      \centerline{\footnotesize{cHKS multiscale BoF}}
\end{minipage}
\end{center}
   \caption{\label{fig:matches1} \small  Retrieval results using different methods. First column: query shapes, second column: first three matches obtained with HKS-based BoF \cite{BroBroOvsGui09}, third column: first three matches obtained using color histograms, fourth column: first three matches obtained with the proposed method (cHKS-based multiscale BoF). Shape annotation follows the convention {\em shapeid.transformation.strength};
   numbers below show distance from query.  Only a single correct match exists in the database (marked in green), and ideally, it should be the first one.}
\end{figure*}

{\bf Results. }
Tables~\ref{tab:BoFS-0}--\ref{tab:multi-resolution-BoFS} summarize the
results of our experiments.
Geometry only descriptor (HKS) \cite{BroBroOvsGui09} is invariant to
photometric transformations, but is somewhat sensitive to topological
noise and missing parts (Table~\ref{tab:BoFS-0}).
On the other hand, the color-only descriptor works well only for
geometric transformations that do not change the shape color.
Photometric transformations, however, make such a descriptor almost
useless (Table~\ref{tab:cBoFH}).
MeshHOG is almost invariant to photometric transformations being based
on texture gradients, but is sensitive to color noise
(Table~\ref{tab:-BOFS-vocab48.mat}).
The fusion of the geometric and photometric data using our approach
(Table~\ref{tab:multi-resolution-BoFS}) achieves nearly perfect
retrieval for mixed and photometric transformations and outperforms
other approaches.
Figure~\ref{fig:matches1} visualizes a few examples of the retrieved
shapes ordered by relevance, which
is inversely proportional to the distance from the query shape.

\begin{table}[tb]
\centering
\begin{tabular}{lccccc}
& \multicolumn{5}{c}{\bf\small Strength} \\
\cline{2-6}
{\bf\small Transform.} & {\bf\small 1} & {\bf\small $\leq$2} & {\bf\small $\leq$3} & {\bf\small $\leq$4} & {\bf\small $\leq$5}\\
\hline
{\small\em Isom+Topo} & {\small 100.00} & {\small 100.00} & {\small 96.67} & {\small 95.00} & {\small 90.00} \\
{\small\em Partial} & {\small 66.67} & {\small 60.42} & {\small 63.89} & {\small 63.28} & {\small 63.63} \\
\hline
{\small\em Contrast} & {\small 100.00} & {\small 100.00} & {\small 100.00} & {\small 100.00} & {\small 100.00} \\
{\small\em Brightness} & {\small 100.00} & {\small 100.00} & {\small 100.00} & {\small 100.00} & {\small 100.00} \\
{\small\em Hue} & {\small 100.00} & {\small 100.00} & {\small 100.00} & {\small 100.00} & {\small 100.00} \\
{\small\em Saturation} & {\small 100.00} & {\small 100.00} & {\small 100.00} & {\small 100.00} & {\small 100.00} \\
{\small\em Noise} & {\small 100.00} & {\small 100.00} & {\small 100.00} & {\small 100.00} & {\small 100.00} \\
\hline
{\small\em Mixed} & {\small 90.00} & {\small 95.00} & {\small 93.33} & {\small 95.00} & {\small 96.00} \\
\end{tabular}
\caption{\small Performance (mAP in \%) of ShapeGoogle using BoFs with HKS descriptors. \label{tab:BoFS-0}}
\end{table}

\begin{table}[tb]
\centering
\begin{tabular}{lccccc}
& \multicolumn{5}{c}{\bf\small Strength} \\
\cline{2-6}
{\bf\small Transform.} & {\bf\small 1} & {\bf\small $\leq$2} & {\bf\small $\leq$3} & {\bf\small $\leq$4} & {\bf\small $\leq$5}\\
\hline
{\small\em Isom+Topo} & {\small 100.00} & {\small 100.00} & {\small 100.00} & {\small 100.00} & {\small 100.00} \\
{\small\em Partial} & {\small 100.00} & {\small 100.00} & {\small 100.00} & {\small 100.00} & {\small 100.00} \\
\hline
{\small\em Contrast} & {\small 100.00} & {\small 90.83} & {\small 80.30} & {\small 71.88} & {\small 63.95} \\
{\small\em Brightness} & {\small 88.33} & {\small 80.56} & {\small 65.56} & {\small 53.21} & {\small 44.81} \\
{\small\em Hue} & {\small 11.35} & {\small 8.38} & {\small 6.81} & {\small 6.05} & {\small 5.49} \\
{\small\em Saturation} & {\small 17.47} & {\small 14.57} & {\small 12.18} & {\small 10.67} & {\small 9.74} \\
{\small\em Noise} & {\small 100.00} & {\small 100.00} & {\small 93.33} & {\small 85.00} & {\small 74.70} \\
\hline
{\small\em Mixed} & {\small 28.07} & {\small 25.99} & {\small 20.31} & {\small 17.62} & {\small 15.38} \\
\end{tabular}
\caption{\small Performance (mAP in \%) of color histograms. \label{tab:cBoFH}}
\end{table}

\begin{table}[!h]
\centering
\begin{tabular}{lccccc}
& \multicolumn{5}{c}{\bf\small Strength} \\
\cline{2-6}
{\bf\small Transform.} & {\bf\small 1} & {\bf\small $\leq$2} & {\bf\small $\leq$3} & {\bf\small $\leq$4} & {\bf\small $\leq$5}\\
\hline
{\small\em Isom+Topo} & {\small 100.00} & {\small 95.00} & {\small 96.67} & {\small 94.17} & {\small 95.33} \\
{\small\em Partial} & {\small 75.00} & {\small 61.15} & {\small 69.93} & {\small 68.28} & {\small 68.79} \\
\hline
{\small\em Contrast} & {\small 100.00} & {\small 100.00} & {\small 100.00} & {\small 98.33} & {\small 94.17} \\
{\small\em Brightness} & {\small 100.00} & {\small 100.00} & {\small 100.00} & {\small 100.00} & {\small 99.00} \\
{\small\em Hue} & {\small 100.00} & {\small 100.00} & {\small 100.00} & {\small 100.00} & {\small 100.00} \\
{\small\em Saturation} & {\small 100.00} & {\small 100.00} & {\small 100.00} & {\small 98.75} & {\small 99.00} \\
{\small\em Noise} & {\small 100.00} & {\small 100.00} & {\small 88.89} & {\small 83.33} & {\small 78.33} \\
\hline
{\small\em Mixed} & {\small 100.00} & {\small 100.00} & {\small 100.00} & {\small 93.33} & {\small 83.40} \\
\end{tabular}
\caption{\small Performance (mAP in \%) of BoFs using MeshHOG descriptors. \label{tab:-BOFS-vocab48.mat}}
\end{table}

\begin{table}[tb]
\centering
\begin{tabular}{lccccc}
& \multicolumn{5}{c}{\bf\small Strength} \\
\cline{2-6}
{\bf\small Transform.} & {\bf\small 1} & {\bf\small $\leq$2} & {\bf\small $\leq$3} & {\bf\small $\leq$4} & {\bf\small $\leq$5}\\
\hline
{\small\em Isom+Topo} & {\small 100.00} & {\small 100.00} & {\small 96.67} & {\small 97.50} & {\small 94.00} \\
{\small\em Partial} & {\small 68.75} & {\small 68.13} & {\small 69.03} & {\small 67.40} & {\small 67.13} \\
\hline
{\small\em Contrast} & {\small 100.00} & {\small 100.00} & {\small 100.00} & {\small 100.00} & {\small 100.00} \\
{\small\em Brightness} & {\small 100.00} & {\small 100.00} & {\small 100.00} & {\small 100.00} & {\small 100.00} \\
{\small\em Hue} & {\small 100.00} & {\small 100.00} & {\small 100.00} & {\small 100.00} & {\small 100.00} \\
{\small\em Saturation} & {\small 100.00} & {\small 100.00} & {\small 100.00} & {\small 100.00} & {\small 100.00} \\
{\small\em Noise} & {\small 100.00} & {\small 100.00} & {\small 100.00} & {\small 100.00} & {\small 100.00} \\
\hline
{\small\em Mixed} & {\small 100.00} & {\small 100.00} & {\small 96.67} & {\small 97.50} & {\small 98.00} \\
\end{tabular}
\caption{\small Performance (mAP in \%) of ShapeGoogle using w-multi-scale BoFs with cHKS descriptors. \label{tab:multi-resolution-BoFS}}
\end{table} 

\subsection{Shape distributions}


{\bf Spectral shape distances. }
Recent works \cite{rustamov2007lbe,MS_GMOD} showed that global shape descriptors can be constructed considering {\em distributions} of intrinsic distances.
Given some intrinsic distance metric $d_X$, 
its cumulative distribution is computed as
\begin{eqnarray}
F_X(\delta) &=& \int \chi_{d_X(x,x') \leq \delta} \, da(x) da(x'),
\end{eqnarray}
where $\chi$ denotes an indicator function.
Given two shapes $X$ and $Y$ with the corresponding distance metrics $d_X, d_Y$, the
similarity (referred to as {\em spectral distance}) is computed as a distance 
between the corresponding distributions $F_X$ and $F_Y$. 

Using the proposed framework, we construct diffusion distances according to~(\ref{eq:diffdist}), where the standard Laplace-Beltrami operator is again replaced by the one associated with the geometric-photometric embedding. Such distances account for photometric information, and, as we show in the following, show superior performance.

{\bf Methods. }
Using the same benchmark as above, we compared shape retrieval approaches that use distance distribution as shape descriptors. Two methods were compared: pure geometric and joint geometric-photometric distances.
In the former, we used average of diffusion distances
\begin{eqnarray}
d(x,x') &=& \frac{1}{|T|}\sum_{t\in T}d_t(x,x'),
\label{eq:spec-dist-photo}
\end{eqnarray}
computed at two scales, $T=\{1024, 4096\}$.
In the latter, the distances were also computed at multiple scales $\eta$ of the photometric component,
\begin{eqnarray}
d(x,x') &=& \frac{1}{|T|}\sum_{t\in T}\prod_{\eta\in H} d_{t,\eta}(x,x').
\label{eq:spec-dist-joint}
\end{eqnarray}
The values $H=\{0,0.1,0.2\}$ were used.
For the computation of distributions, the shapes were subsampled at $2500$ points using the farthest point sampling algorithm.

{\bf Results. }
Tables~\ref{tab:regular:diff}--\ref{tab:mltsc-sum} summarize the results.
Both descriptors appear insensitive to photometric transformations. The joint distance has superior performance in pure geometric and mixed transformations.
We conclude that the use of non-zero weight for the color component adds discriminativity to the distance distribution descriptor, while being still robust under photometric transformations.

\begin{table}[tb]
\centering
\begin{tabular}{lccccc}
& \multicolumn{5}{c}{\bf\small Strength} \\
\cline{2-6}
{\bf\small Transform.}& {\bf\small 1} & {\bf\small $\leq$2} & {\bf\small $\leq$3} & {\bf\small $\leq$4} & {\bf\small $\leq$5}\\
\hline
{\small\em Isom+Topo} & {\small 80.00} & {\small 90.00} & {\small 88.89} & {\small 86.67} & {\small 89.33} \\
{\small\em Partial} & {\small 56.25} & {\small 65.62} & {\small 61.61} & {\small 58.71} & {\small 61.13} \\
\hline
{\small\em Contrast} & {\small 100.00} & {\small 100.00} & {\small 100.00} & {\small 100.00} & {\small 100.00} \\
{\small\em Brightness} & {\small 100.00} & {\small 100.00} & {\small 100.00} & {\small 100.00} & {\small 100.00} \\
{\small\em Hue} & {\small 100.00} & {\small 100.00} & {\small 100.00} & {\small 100.00} & {\small 100.00} \\
{\small\em Saturation} & {\small 100.00} & {\small 100.00} & {\small 100.00} & {\small 100.00} & {\small 100.00} \\
{\small\em Noise} & {\small 100.00} & {\small 100.00} & {\small 100.00} & {\small 100.00} & {\small 100.00} \\
\hline
{\small\em Mixed} & {\small 66.67} & {\small 73.33} & {\small 78.89} & {\small 81.67} & {\small 81.33} \\
\end{tabular}
\caption{\small Performance (mAP in \%) of pure geometric spectral shape distance (\ref{eq:spec-dist-photo}). \label{tab:regular:diff}}
\end{table} 

\begin{table}[tb]
\centering
\begin{tabular}{lccccc}
& \multicolumn{5}{c}{\bf\small Strength} \\
\cline{2-6}
{\bf\small Transform.} & {\bf\small 1} & {\bf\small $\leq$2} & {\bf\small $\leq$3} & {\bf\small $\leq$4} & {\bf\small $\leq$5}\\
\hline
{\small\em Isom+Topo} & {\small 100.00} & {\small 100.00} & {\small 100.00} & {\small 100.00} & {\small 100.00} \\
{\small\em Partial} & {\small 62.50} & {\small 72.92} & {\small 65.97} & {\small 62.50} & {\small 67.50} \\
\hline
{\small\em Contrast} & {\small 100.00} & {\small 100.00} & {\small 100.00} & {\small 100.00} & {\small 100.00} \\
{\small\em Brightness} & {\small 100.00} & {\small 100.00} & {\small 100.00} & {\small 100.00} & {\small 100.00} \\
{\small\em Hue} & {\small 100.00} & {\small 100.00} & {\small 100.00} & {\small 100.00} & {\small 100.00} \\
{\small\em Saturation} & {\small 100.00} & {\small 100.00} & {\small 100.00} & {\small 100.00} & {\small 100.00} \\
{\small\em Noise} & {\small 100.00} & {\small 100.00} & {\small 100.00} & {\small 100.00} & {\small 100.00} \\
\hline
{\small\em Mixed} & {\small 100.00} & {\small 93.33} & {\small 95.56} & {\small 96.67} & {\small 93.70} \\
\end{tabular}
\caption{\small Performance of (mAP in \%) of the multiscale joint geometric-photometric spectral distance (\ref{eq:spec-dist-joint}). \label{tab:mltsc-sum}}
\end{table}  
\section{Conclusions}
\label{sec:concl}

In this paper, we explored a way to fuse geometric and photometric information in the construction of shape descriptors. Our approach is based on heat propagation on a manifold embedded into a combined geometry-color space. Such diffusion processes capture both geometric and photometric information and give rise to local and global diffusion geometry (heat kernels and diffusion distances), which can be used as informative shape descriptors. 
We showed experimentally that the proposed descriptors outperform other geometry-only and photometry-only descriptors, as well as state-of-the-art joint geometric-photometric descriptors.
In the future, it would be important to formally characterize the isometry group induced by the joint metric in order to understand the invariant properties of the proposed diffusion geometry, and possibly design application-specific invariant descriptors. 
%

%



\bibliographystyle{plain}
\bibliography{bib_art}

\end{document}